\documentclass[3p,times]{elsarticle}
\usepackage{ecrc}

\makeatletter
\def\ps@pprintTitle{%
 \let\@oddhead\@empty
 \let\@evenhead\@empty
 \def\@oddfoot{}%
 \let\@evenfoot\@oddfoot}
\makeatother
\volume{00}

\firstpage{1}

\runauth{Faisal R. Al-Osaimi}

\usepackage{amssymb}
\usepackage{mathrsfs}
\usepackage{amsmath}

\usepackage[figuresright]{rotating}

\usepackage{setspace}  %%%*
\begin{document}

\begin{frontmatter}



\dochead{}
%% e.g. \dochead{17th International Conference on Dynamical Processes in Excited States of Solids}

\title{Learning Descriptors Invariance Through Equivalence Relations Within Manifold: A New Approach to Expression Invariant 3D Face Recognition}


\author{Faisal R. Al-Osaimi}
\address{The Department of Computer Engineering,
Umm Al-Qura University,
Makkah, Saudi Arabia. E-mail: frosaimi@uqu.edu.sa}

%\address{}

\begin{abstract}

This paper presents a unique approach for the dichotomy between useful and adverse variations of key-point descriptors, namely the identity and the expression variations in the descriptor (feature) space. The descriptors variations are learned from training examples. Based on the labels of the training data, the equivalence relations among the descriptors are established.  
Both types of descriptor variations are represented by a graph embedded in the descriptor manifold. The invariant recognition is then conducted as a graph search problem. A heuristic graph search algorithm suitable for the recognition under this setup was devised. The proposed approach was tests on the FRGC v2.0, the Bosphorus and the 3D TEC datasets. It has shown to enhance the recognition performance, under expression variations in particular, by considerable margins.

\end{abstract}


\end{frontmatter}

%%\linenumbers 

\section{Introduction}
\label{sec_intro}

The 3D face recognition has shown to achieve considerable recognition accuracy and robustness, especially when compared to its 2D counterpart. There are vital applications of face recognition such as security, access control and human-machine interaction.  The importance of its applications combined with the recent advances in the 3D digitization technologies has been the driving force behind the noticeable interest in 3D face recognition among the researchers in the computer vision and pattern recognition community. Despite the reported advances in 3D face recognition in the recent years, the practical applications of 3D face recognition require even higher accuracies and robustness.      

A particularly interesting recognition paradigm is that concerned with first the detection of key-points and then the description of the local 3D surface around them. Approaches based on this paradigm inherently enjoys desirable properties such as the robustness to clutter and occlusions and the enabling of partial surface matching. While this paradigm has been very successful in the general object recognition \cite{lowe1999object,belongie2002shape,rublee2011orb}, the face recognition performance of the well-known approaches in this paradigm remained, until recently, below that of the state-of-the-art (e.g. \cite{mian2008keypoint,al2007interest}). Recently, the rotation-invariant and adjustable integral kernel (RAIK) approach \cite{al2015novel} which belongs to this paradigm has shown to highly perform in 3D face recognition when the matching is limited to the semi-rigid regions of the face (the nose and the forehead). The RAIK descriptors are discriminative, invariant to the 3D rotations and completely representative of the underling surface. In the context of 3D face recognition, the descriptors also encode the expression variations. %ing its variations (i.e., the expression variations in the context of 3D face recognition). 
For this reason, the RAIK descriptor was deemed an appropriate choice for the proposed work. 
The 3D face variations pertaining to the identities of the individuals are essential for correct recognition. However, in practice these valuable variations mix with other types of variations that may drawback the performance and the robustness of the face recognition. These adverse variations arise basically from the challenging expression deformations of the facial surface and its rigid transformations. %While invariant recognition to the rigid transformations is already achievable, expression invariance remains more challenging to tackle. 

As the RAIK descriptors faithfully encode the shape information of the local surface independently from the rigid transformations (including surface and 3D scanner orientations), it is imperative to assume that the local descriptors represent only the identity related shapes and the expression deformations of the local surfaces. One can consider the expression deformations to {\em displace} the descriptors, in the descriptor space, in an unknown and complex way. %which depends on the undeformed local surface, deformation and the descriptor itself. 
As any descriptor matching approach will define a similarity (or a dissimilarity) measure based on their locations in the descriptor space, the expression-deformation-induced displacements are likely to adversely affect the recognition performance. In the proposed work, the differentiation between the two types of the descriptor displacements, namely the identity-induced displacements (IIDs) and the expression-induced displacements (EIDs), is learned and utilized to enhance the recognition. This constitutes the main contribution of the proposed work. Knowing that any two displaced training descriptors came from the same key-point (i.e., from the same location on the face) and the same person, but only different due to the expression deformation, provides a piece of information. The consolidation of such pieces of information can be useful in the differentiation between the two types of displacements for unseen descriptors and 3D faces. This information along with the displacements that set different individuals apart is represented by a large graph embedded in the descriptor manifold for the utilization in the expression invariant face recognition. It should be noted that the aforementioned displacements can be also perceived as one-to-one relations from the descriptor space to itself. In this terminology, the sub-graphs corresponding to the descriptors (the nodes) of the same individual under varying facial expressions represent equivalence relations (ERs) but the sub-graphs connecting the descriptors of different individuals represent identity relations (IRs).

The rest of the paper is organized as follows. An overview of the proposed approach is provided in section \ref{sec_ov}. Next, the relevant literature is reviewed in section \ref{sec_lit}.  The proposed approach is then detailed in section \ref{sec_app}. The conducted experiments and their results are described and discussed in section \ref{sec_exp}.

\section{An Overview of the Proposed Approach}
\label{sec_ov}

\begin{figure*}[!t]
\includegraphics[width=6.4in]{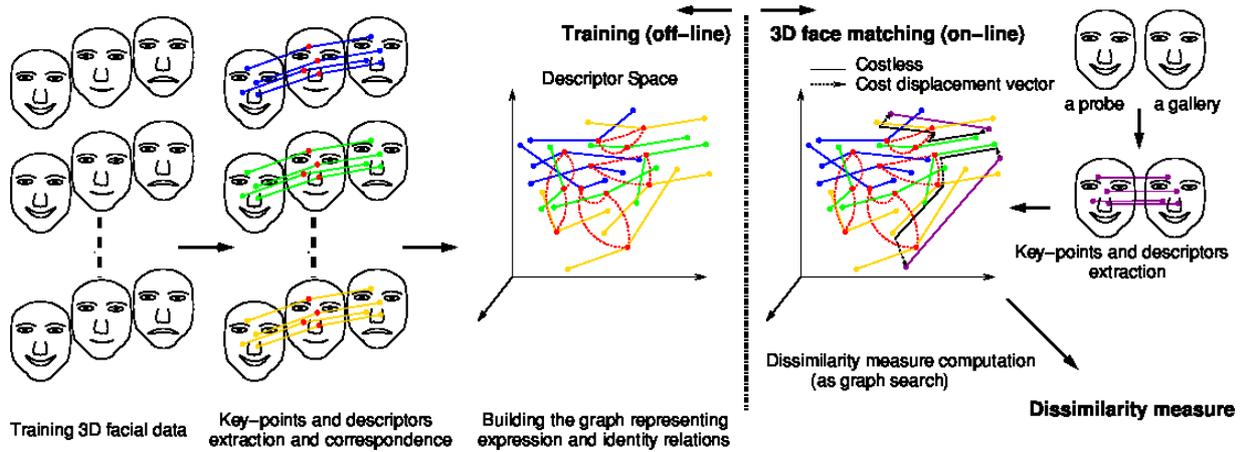}
\caption{An Overview illustration of the proposed approach (better seen in color).}
\label{fig_ov}       % Give a unique label
\end{figure*}

Before giving an overview description of the proposed approach, the definitions of some terms used in the paper are provided. The set of the RAIK descriptors that are extracted from a 3D facial scan is called an {\em ensemble}. The set of ensembles that are extracted from multiple 3D facial scans of a person is call a {\em collection}. %The set of collections are referred to as {\em aggregate}, i.e., an aggregate contains collections where each collection belong to a unique individual.

In the proposed approach, many training collections of the RAIK descriptors are extracted from sets of 3D facial scans of many individuals. The 3D facial scans in each collection are under varying expressions, including the neutral expression. The identity of the collections are unique, i.e, there are no more than one collection that belong to the same individual. The descriptors within each collection are then corresponded. In this correspondence, the descriptors of each ensemble are mapped (linked) to their corresponding descriptors in the other ensembles. This results in multiple sets of corresponded descriptors. Each corresponded set of the descriptors (within each collection) is then joined together by a simple spanning graph which represents the equivalence relations, ERs.    
%Instead of fully connecting the corresponding set of descriptors (within a collection), the simplest spanning graph (a line joining the descriptors) is sufficiently used to represent the equivalence relations, ERs. There are an many ERs in each collection as the number of descriptors in each ensemble. 
Between the different collections, the corresponding equivalence graphs (ERs) are connected to each other, enabling the representation of the identity relations, IRs. One viable approach is to inter-connect the different equivalence graphs from the descriptors of a neutral ensemble to the descriptors of a neutral ensemble of the other collections (neutral to neutral connections). All the previous computations take place during an off-line training phase.    
%The connection of the different ERs is made from the descriptors of in a neutral ensemble in one collection to the descriptors in a neutral ensemble in another collection. This choice is logical and viable for the representation of IRs in the graph. All previous computations take place during an off-line training phase. 

For the invariant matching of a probe ensemble to a gallery ensemble (the on-line matching phase), the descriptors are first corresponded. Then, for each corresponded pair of descriptors, a dissimilarity measure is computed. The dissimilarity measures are eventually combined in an overall dissimilarity measure. For the computation of the dissimilarity measures, the graph is searched for the path connecting each corresponded pair such that; the encountered IRs along the graph search path have cost (as a vector quantity), the encountered ERs have zero-cost and the magnitude of the sum of all the encountered IRs is minimum. Finally, on the basis of the minimized IR quantities the dissimilarity measures are computed. See fig. \ref{fig_ov} for an overview graphical illustration.     

\section{The Literature Review}
\label{sec_lit}

The approaches to the 3D expression invariant face recognition can be broadly categorized into two categories; the image space approaches and the feature space approaches. As dealing with 3D surfaces in the image space is more intuitive, it is not surprising to find that the early successful approaches are in this category. One class of the image space methods is based on the avoidance of the highly deformable regions of the face such as the mouth and the cheeks, e.g., \cite{mian_2007,chang_2005,chang2006multiple,faltemier2008region}, which has shown to be considerably effective. Unfortunately, the considered regions of the face still undergo expression deformations which can adversely affect the recognition accuracy. Another class of methods in the same category is to deform the 3D facial surface in an attempt to remove or at least reduce the effects of the expression deformations. The approach proposed in \cite{bronstein2003expression}, transforms the 3D facial surface by bending the surface while preserving the geodesic distances among its points into an invariant image. As that approach flattens the expressions, it also flattens some other facial surface geometries that can be important to the recognition. Recently, several methods extract expression invariant measures from geodesic distances of the facial surface \cite{drira20133d,smeets2012isometric,berretti20103d}. The annotated face model, which is an elastic deformable model \cite{metaxas2002elastically}, is used deform a 3D face under one expression to a target 3D face under another expression \cite{murtuza2007three}. The recognition is then performed on the fitted model. It is not clear how such model can differentiate between the expression deformations and the other unique geometries of the face which are important for the recognition. In another work \cite{al2009expression}, a low dimensional principal component analysis (PCA) subspace in which the expression deformations reside is found from the 3D shape residues of registered pairs of non-neutral and neutral scans, where each pair belong to the same subject. For an invariant recognition, the expression deformations in the image space are separated from novel 3D residues based on their projection onto the subspace and then their reconstruction. These methods are better situated for global face matching. Therefore, their robustness and accuracy may be undermined under occlusions and wide pose variations. The work in \cite{maalej2011shape}, performs elastic invariant matching of local surface regions around a handful of manually picked facial anatomical landmarks. Nonetheless, reliable automatic localization of the facial landmarks remains an open problem.

In a wide range of approaches, the matching is performed based on extracted features, e.g., \cite{gokberk20063d,lu2005multimodal,xu2009automatic,tan2007fusing,wang2010robust,singh2011face}. Feature extraction reduces the dimensionality of the facial data. An independent set of features is preferable for the recognition. Some features may be less sensitive to the expressions than others, providing a limited level of expression invariance. Typical examples of feature extraction methods are the linear subspace methods such as the different variants of PCA \cite{turk1991eigenfaces,belhumeur1997eigenfaces,blanz2003face,wang2004unified,russ20063d}, linear discriminant analysis (LDA) \cite{wang2004unified,belhumeur1997eigenfaces} and independent component analysis (ICA) \cite{draper2003recognizing,harguess2009case,liu2003independent}. %Alignment is crucial for these methods.

In the feature space, there are different methods in the literature that attempt to find the space regions that are associated with certain identities or a class labels. One widely used methods is to estimate a probability density function (PDF) over the space regions. The estimation of a PDF in a multidimensional vector space faces practical challenges. The number of training data samples required to compute such PDF grows exponentially with the number of dimensions, referred to as ``the curse of dimensionality" \cite{verleysen2005curse,jain1997feature}. Strong assumptions about the PDF are often made, such as assuming normal distribution \cite{paalanen2006feature,raudys1991small}, to survive on the available training samples. The ``kernel trick" method was used in several approaches feature space approaches, e.g., \cite{scholkopf1998nonlinear,hotta2008robust,ksantini2011new}.  The kernel function replaces the dot product in their non-kernelized variants, e.g., the support vector machines k-SVMs \cite{ksantini2011new} and k-PCA \cite{yang2002kernel}, they are often data driven. Its overall process can be considered as a nonlinear transformation of the feature space to induce more separable classes. While it provides an elegant approach for the non-linear separability, it does not explicitly address the expression variations which may render the facial classes inseparable. In contrast, the proposed work is explicit in handling the expression variations and versatile in selecting the relevant subset of the driving data for each match.

The manifold methods \cite{balasubramanian2007biased,chang2006manifold,zhang2007linear,lin2006riemannian,pan2009weighted} in most cases are concerned with the non-linear dimensionality reduction, where the feature distribution in the feature space may be locally of a much lower dimension than that of the manifold as a whole. The de facto example is a ``Swiss roll'' surface residing in a higher dimensional space. Typically, rather than using the Euclidean distances for matching, the geodesic distances on the manifold, e.g., \cite{samir2006three} are used instead. Sometimes the problem at consideration or its formulation guarantees that the feature data form a manifold of a specific low dimension and the availability of enough feature samples to recover the lower dimensional manifold. An example of such a problem is the manifold of a rotated template image \cite{raytchev2004head}, the manifold dimension in this case is the number of the degree of freedom and the data samples can be generated as needed. The expression and identity manifold is of a complex structure with several dimensions (possibly variable locally).  This problem also lacks the availability of enough data samples to accurately recover the manifold. In contrast, the proposed does not attempt to recover the lower dimension of the manifold, unroll it, or extract geodesic distances but rather the manifold is perceived as a sparse distribution of the data samples and the distances between certain sparse points are shortened to zero.

\section{The Proposed Approach}
\label{sec_app}

This section describes the steps of the proposed approach and discusses the concepts behind them in more details than previously provided in the introduction and the overview sections, Section \ref{sec_intro} and \ref{sec_ov}.  

\subsection{Conceptual Analysis}
\label{sec_ca}

A manifold has the notion of the Euclidean spaces (tangential spaces) locally at each of its points. Conventionally, a descriptor space is either treated as one Euclidean space or as a manifold. % with a strong notion of locally Euclidean spaces. 
In both situations, the large distances in the descriptor space translate into large dissimilarity measures which is not plausible for the recognition under variations, e.g., the 3D face recognition under expression variations.
In contrast, the proposed approach has the ability to bring and merge distant tangential spaces of the manifold with each other. %with the local tangential space of the manifold. 
Consequently, the proximity and the non-proximity among the manifold points can combine and contribute more meaningfully to the dissimilarity measure.    
%the proposed approach combines the notion of descriptor manifold locality (the proximity) with the ability to bring and merge distant tangential spaces (the non-locality) with the local tangential space. 
% to the non-locality of the descriptor manifold. 
%In contrast, the proposed approach combines manifold locality (proximity) of the descriptor manifold, not to be confused with the locality of the RAIK descriptors in the image space, with the non-locality of the descriptor manifold. 
This can be achieved through the establishment of equivalence relations between reference points (corresponded descriptors) in the descriptor manifold. Let $\mathcal Q = \{D_1, \dots,D_n\}$ be a set of corresponded descriptors (an equivalence set) and $T_{D_i}M$ denotes a tangential space at the $i$-th descriptor of the manifold, $M$. Each corresponded pair of descriptors, $\{D_i, D_j\}$, establishes an equivalence between all the tangential spaces in which $D_i$ and  $D_j$ exist, i.e., the equivalence also extends to the manifold points around the descriptors. The same concept similarly applies for all the pair combinations of the descriptors in $\mathcal Q$. This gives rise to the notion of the tangential space as quotient space, $Q_{\mathcal Q}M$. 

\begin{figure}[!t]
\begin{tabular}{c c}
\centering
      \includegraphics[width=0.4\columnwidth]{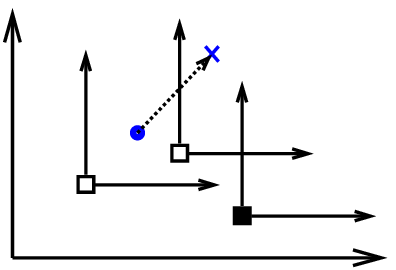} & \includegraphics[width=0.4\columnwidth]{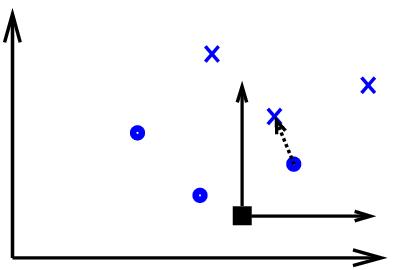} \\
      {\small (a)} &  {\small (b)} 
\end{tabular}
\caption{In (a), two manifold points apprearing in the tangential spaces at three equivalent manifold points (the square ones). The displacement vector between the two points is shown. In (b), the orignal two points and their images under the equivalency are shown with respect to the reference tangential space (the solid square). The invariant displacement vector under the equivaleny is also shown.}
\label{fig_inv_disp}       % Give a unique label
\end{figure}

Distance and vector quantities, namely the displacement vectors based on which the dissimilarity measures are computed, can be computed in the quotient space. Let $\bf x$ and $\bf y$ be two manifold points and the displacement between them ${\bf d}({\bf x}, {\bf y})$ is to be computed under the equivalence set $\mathcal Q$.
%, $T_{D_{1,\dots,n}}M$.
% to be computed in all the tangential spaces at the equivalence set $\mathcal Q$, $T_{D_{1,\dots,n}}M$. 
%To compute the distance between $\bf x$ and $\bf y$, 
The equivalent images of $\bf x$ and $\bf y$ in all the tangential spaces at the different $\mathcal Q$ points are mapped to a reference tangential space (the quotient space) which can be the tangential space at any $\mathcal Q$ point. The mapping from the $i$-th tangential space to the $r$-th tangential space, ${\mathcal E}_i^{r}({\bf x})$, is provided by Eqn. \ref{eqv_map1} and similarly the mapping ${\mathcal E}_j^{r}({\bf x})$ is provided by Eqn. \ref{eqv_map2}. In the reference space there will be multiple images of $\bf x$ and $\bf y$ points and the displacement between them is defined as the displacement with the minimum magnitude (norm) between any equivalent image of $x$ to any equivalent image of $\bf y$, Eqn. \ref{eq_disp}, where $\|.\|$ is the vector norm of the displacement vector and ${\bf t} = (t_1, t_2, t_3)$ is the optimal triple of equivalent descriptors. A graphical illustration is provided in Fig. \ref{fig_inv_disp}.
\begin{eqnarray}
{\mathcal E}_i^{r}({\bf x}) &=&  {\bf x} + D_r- D_i.    \label{eqv_map1}\\
{\mathcal E}_j^{r}({\bf y}) &=&  {\bf y} + D_r- D_j.    \label{eqv_map2}\\
{\bf t} &=&  {\arg\min}_{(D_i, D_j, D_r) \subset {\mathcal Q^{3}}} \| {\mathcal E}_i^r({\bf x}) - {\mathcal E}_j^r({\bf y}) \|. \label{eq_disp} \\
{\bf d}({\bf x}, {\bf y}) &=&  {\mathcal E}_{t_1}^{t_3}({\bf x}) - {\mathcal E}_{t_2}^{t_3}({\bf y}). \label{eq_disp}
\end{eqnarray}

On the basis of the mapping relations, Eqns \ref{eqv_map1} and \ref{eqv_map2}, the differentiation between the expression variations and the identity variations can be achieved. The expression variation of a descriptor from one (the $i$-th) expression to another (the $j$-th) can be considered as the additive displacement vector ${\Delta}{\mathcal E}_{i}^{j} = D_j - D_i$ and then separated from the identity variation. Therefore, ${\mathcal E}_i^{j}({\bf .})$ modifies (or displaces) the expression of the input descriptor from expression $i$ to expression $j$. 

%Let the identity of the equivalence set $\mathcal Q$ be the space point $\bf c$. 
In the next discussions, the term ``identity'' refers to the person identity for the collections and the ensembles (the scans as well) but to a space point (location) for the descriptors and the equivalence sets, i.e., their typical unexpressed space location for the specified person. Suppose that the identity of the descriptor $\bf x$ is ${\bf p}_1$ and the identity of the descriptor $\bf y$ is ${\bf p}_2$. The descriptors $\bf x$ and $\bf y$ have both expression (${\bf e}_x$ and ${\bf e}_y$) and identity (${\mathcal I}({\bf x})$ and ${\mathcal I}({\bf y})$) components, Eqns \ref{eq_desc_decomp1} and \ref{eq_desc_decomp2}. The identity variation ${\Delta}{\mathcal I}_{p_1}^{p_2}({\bf x}, {\bf y})$ is defined as the difference between the two identity components, Eqn \ref{eq_id_var}.  
\begin{eqnarray}
{\bf x} &=& {\mathcal I}({\bf x})+ {\bf e}_x.    \label{eq_desc_decomp1}\\
{\bf y} &=& {\mathcal I}({\bf y})+ {\bf e}_y.    \label{eq_desc_decomp2}\\
{\Delta}{\mathcal I}_{p_1}^{p_2}({\bf x}, {\bf y}) &=& {\mathcal I}({\bf y}) - {\mathcal I}({\bf x}).    \label{eq_id_var}
\end{eqnarray}
The identity variation ${\Delta}{\mathcal I}_{p_1}^{p_2}({\bf x}, {\bf y})$ can be computed based on the displacement ${\bf d}({\bf x}, {\bf y})$ (Eqn \ref{eq_disp}) as shown in Eqns. \ref{eq_comp_i4} and \ref{eq_comp_i5}.       
\begin{eqnarray}
{\bf d}({\bf x}, {\bf y}) &=& {\mathcal E}_i^r({\bf x}) - {\mathcal E}_j^r({\bf y}),    \label{eq_comp_i1}\\
                           &=& {\bf x} - {\bf y} + D_j- D_i,      \label{eq_comp_i2}\\
                           &=& {\mathcal I}({\bf x})+ {\bf e}_x - {\mathcal I}({\bf y}) - {\bf e}_y + D_j- D_i, \; \mbox{and}    \label{eq_comp_i3}\\
                           &=& - {\Delta}{\mathcal I}_{p_1}^{p_2}({\bf x}, {\bf y}) - \Delta {\mathcal E}_x^i({\bf x}) + \Delta{\mathcal E}_y^j({\bf y}).    \label{eq_comp_i4}
\end{eqnarray}
When the expression of ${\bf x}$ is close (or ideally equals to) the expression $D_i$ and that of ${\bf y}$ equals to the expression $D_j$ then ${\Delta}{\mathcal I}_{p_1}^{p_2}$ reduces to $-{\bf d}({\bf x}, {\bf y})$, Eqn \ref{eq_comp_i5}. This requirement is seamlessly achieved during the computation of ${\bf d}({\bf x}, {\bf y})$ according to Eqn. \ref{eq_disp}. 
\begin{eqnarray}
{\Delta}{\mathcal I}_{p_1}^{p_2}({\bf x}, {\bf y}) = - {\bf d}({\bf x}, {\bf y}).  \label{eq_comp_i5}
\end{eqnarray}
While the identity variations $\Delta{\mathcal I}_{p_1}^{p_2}({\bf x}, {\bf y})$ is computed based on the descriptors ${\bf x}$ and ${\bf y}$, $\Delta{\mathcal I}_{p_1}^{p_2}(., .)$ is independent of any particular descriptors as long they belong to the specified identities and at the corresponding key-points on the face. Therefore, the norm value of $\Delta{\mathcal I}_{p_1}^{p_2}({\bf x}, {\bf y})$ or equally that of ${\bf d}({\bf x}, {\bf y})$ can be used in the computation of an invariant dissimilarity measure.  

In general, the identity of a descriptor ${\bf x}$ can be changed from ${\bf p}_1$ to the identity of any known descriptor, ${\mathcal I}(.) = {\bf p}_k$, as in Eqn. \ref{eq_ch_i}. 
\begin{eqnarray}
{\mathcal I}_{p_1}^{p_2}({\bf x}) = {\bf x} + {\Delta}{\mathcal I}_{p_1}^{p_k} \Rightarrow {\mathcal I}({\mathcal I}_{p_1}^{p_k}({\bf x})) = {\bf p}_{k} \label{eq_ch_i}
\end{eqnarray}
Similarly, the expression of a descriptor ${\bf x}$ can be changed from ${\bf e}_x$ to that of any known descriptor, ${\mathcal E}(.) = {\bf e}_k$, as in Eqn. \ref{eq_ch_e}.
\begin{eqnarray}
{\mathcal E}_{e_x}^{e_k}({\bf x}) = {\bf x} + {\Delta}{\mathcal E}_{e_x}^{e_k} \Rightarrow {\mathcal E}({\mathcal E}_{e_x}^{e_k}({\bf x})) = {\bf e}_k  \label{eq_ch_e}
\end{eqnarray}
The identity and expression (changing) relations can be composed in any arbitrary sequence, e.g., that given in Eqn. \ref{eq_comp_rw}.  
\begin{eqnarray}
{\bf x}' &=& {\mathcal E}_{e_1}^{e_2} \circ {\mathcal I}_{p_2}^{p_3} \circ {\mathcal E}_{e_x}^{e_1} \circ {\mathcal I}_{p_1}^{p_2} ({\bf x}), \label{eq_comp_rw} \\
         &=& {\mathcal E}_{e_1}^{e_2} \circ {\mathcal E}_{e_x}^{e_1} \circ {\mathcal I}_{p_2}^{p_3} \circ {\mathcal I}_{p_1}^{p_2} ({\bf x}), \mbox{and} \\
         &=& {\mathcal E}_{e_x}^{e_2} \circ {\mathcal I}_{p_1}^{p_3} ({\bf x}). \\
 {\mathcal E}({\bf x}') &=& {\bf e}_2. \\
 {\mathcal I}({\bf x}') &=& {\bf p}_3.         
\end{eqnarray}
The identity and expression relation composition provides the proposed approach with the capacity to utilize the training data in the invariant recognition. Let $D_a$ and $D_b$ be two descriptors for which the invariant dissimilarity measure will be computed. The identity of $D_a$ will be changed to that of an equivalence set in a training collection. Then, within the equivalence set, the expression under the new identity is varied. The process of changing identity and expression is repeated and until eventually arriving to $D_b$. For a suitable choice of the composition sequence, the expression variations will cancel out and the value $\|\Delta{\mathcal I}_{p_a}^{p_b}\|$ will be ideally zero (practically minimal) if ${\bf p}_a = {\bf p}_b$  (i.e., both have the same identity) or non-zero (practically non-minimal) otherwise. The details are provided in Section \ref{sec_gs}. 
%identity and expression of descriptors are varied as ensemble   

%\subsection{Overview of the RAIK Descriptors}

\subsection{The Correspondence of the Descriptors}
\label{sec_cor}

The correspondence of the descriptors is important for the proposed approach, since it enables the tracking of their identity and expression variations. For a pair of descriptor ensembles, ${\mathcal N}_1$ and ${\mathcal N}_2$, the correspondence is defined as the mapping of each descriptor $D_{i} \in {\mathcal N}_1$ to one descriptor $D_{j} \in {\mathcal N}_2$, such that each pair $\{D_i , D_j\}$ correspond to a particular spatial location on the face. %, i.e., in the image space. 
Let the correspondence of ${\mathcal N}_1$ and ${\mathcal N}_2$ be denoted by the set $\mathscr M_{1,2}$ for which Eqn. \ref{eq_dcor} holds true. 
\begin{eqnarray}
{\mathscr M}_{1,2} \subset \{\{D_i, D_j\} \;\; | \;\; D_i \in {\mathcal N}_1 \wedge D_j \in {\mathcal N}_2 \}.  \label{eq_dcor}
\end{eqnarray}

The correspondence is established based on a customized variant of the random sample consensus (RANSAC) algorithm \cite{fischler1981random}. The dissimilarity of the descriptors and their locations are both utilized. 
The descriptors in the two ensembles are initially matched against each other based on the dissimilarity measure in Eqn. \ref{eq_l1}, where the summation is over all the descriptor elements.  
The descriptors in ${\mathcal N}_1$ are corresponded to those in ${\mathcal N}_2$ with the minimum dissimilarity measures.
%$l_1$ norm distance between the descriptors, Eqn. \ref{eq_l1}. The descriptors in ${\mathcal N}_1$ are corresponded to those in ${\mathcal N}_2$ with the minimum $l_1$ distances. 
\begin{eqnarray}
{\bf d}(D_i, D_j) = \sum \|D_i - D_j\|.  \label{eq_l1}
\end{eqnarray}
This will result in many correctly corresponded pairs but also will result in mis-correspondences. Let the key-point sets of the ensembles ${\mathcal N}_1$ and ${\mathcal N}_2$ be denoted respectively by ${\mathscr K}_1$ and ${\mathscr K}_2$, where each key-point ${\bf k}$ in ${\mathscr K}_1$ or ${\mathscr K}_2$ is a vector of the 3D point coordinates, ${\bf k} = [x, y, z]^\top$. To correct the mis-correspondences, the 3D rigid transformations, both the rotation $\bf R$ and the translation ${\bf t}$, that transform the key-points in ${\mathscr K}_1$ to their corresponded ones in ${\mathscr  K}_1$ are first estimated using least square fitting. Then, the error of transformation (the Euclidean distance error), defined in Eqn. \ref{eq_err_tr}, is used to split the correspondences into inliers and outliers by comparison against a threshold value of the transformation error, $e_{th}$. 
\begin{eqnarray}
{\bf k}_i' &= {\bf R}{\bf k}_i + {\bf t}, \;\;\;\;\;\;\;\; & {\bf k}_i \in {\mathscr K}_1. \\
e &= \|{\bf k}_i' - {\bf k}_j\|, \;\;\;\;\;\;\; &  {\bf k}_j \in {\mathscr K}_2. \label{eq_err_tr}
\end{eqnarray}
The rigid transformations are then recalculated on the basis of the inlier correspondences and the process is iterated until the change in the norms of the new $\bf R$ and $\bf t$ diminishes. Using the converged values of $\bf R$ and $\bf t$, the key-points in ${\mathscr K}_1$ are transformed and then re-corresponded only to the ${\mathscr K}_2$ key-points in their vicinity, based on the minimum values of the dissimilarity measure in Eqn. \ref{eq_l1}. %See Fig. \ref{} for an example of corresponded descriptors (or key-points). 

\subsection{The Construction of the Embedded Graph}
\label{sec_cg}

As previously mentioned, the graph is constructed from a training set of descriptors $\mathscr D$ that is organized in a set of ensembles ${\mathscr N}$ and in a set collections $\mathscr C$, too. 
\begin{eqnarray}
\mathscr D &=& \{ D_1, \dots, D_{|\mathcal D|} \}. \\
\mathscr N &=& \{ \mathcal N_1, \dots, \mathcal N_{|\mathcal N|} \}, \;\;\mbox{where}\;\: \forall_{D \in \mathcal N , \; \mathcal N \in \mathscr N} D \in \mathscr D. \\
\mathscr C &=& \{ \mathcal C_1, \dots, \mathcal C_{|\mathcal C|} \}, \;\;\;\;\; \mbox{where}\;\: \forall_{\mathcal N \in \mathcal C, \; \mathcal C \in \mathscr C} \mathcal N \in \mathscr N.
\end{eqnarray}
%(contains the descriptors of a facial scan, $D_i \in {\mathcal N}_j$), collections ${\mathcal C}_{1,\dots,C}$ (contains the ensembles of one person, ${\mathcal N}_k \in {\mathcal C}_l$) and an agregate set ${\mathscr A}$ that containing all training collections (${\mathcal C}_o \in{\mathcal A}$). 
The ERs are extracted from the correspondences within the individual collections, between pairs of ensembles, while the IRs are extracted from those between pairs of collections.
%As previously mentioned, the graph is constructed from training data organized in ensembles ${\mathscr N}_{1,\dots,N}$ (contains the descriptors of a facial scan, $D_i \in {\mathscr N}_j$), collections ${\mathscr C}_{1,\dots,C}$ (contains the ensembles of one person, ${\mathscr N}_k \in {\mathscr C}_l$) and an agregate set ${\mathscr A}$ that containing all training collections (${\mathscr C}_o \in{\mathscr A}$). The ERs are extracted from the correspondences within individual collections, between pairs of ensembles, while IRs are extracted between pairs of collections. 

Initially, sub-graphs representing the ERs are sequentially extracted from each collection ${\mathcal C}_i$ in ${\mathscr C}$. To achieve that, a list of ensemble pair combinations within the collection ${\mathscr C}_i$ is generated and then each generated ensemble pair is corresponded as described in Subsection \ref{sec_cor}. The use of all possible combinations may be overly redundant, especially for large number of ensembles within the collection. This is because the correspondence problem is ideally transitive. The minimum number of such pairs required for connected sub-graphs is $|\mathcal C_i|-1$, where $|\mathcal C_i|$ is the number of ensembles in the collection. However, for robustness a sufficient level of redundancy is maintained, since the redundancy could mitigate possible errors in key-point localization and detection under the different facial expressions. 

A disciplined approach for the maintenance of the required redundancy is based on the reduction of the hope count between the ensembles, when representing the ensembles as graph nodes and the pair combinations as graph edges. First, a minimal connected graph, $\mathscr G_i = (\mathscr V_i, \mathscr E_i)$, is found, e.g., by connecting each ensemble to the next. Then, iteratively the graph diameter, the largest path between any two ensembles, is computed and an edge (an ensemble pair) is introduced between the two ensembles until the graph diameter falls below some predefined threshold. The correspondences are then computed for the ensemble pairs adjacent to every edge in ${\mathscr E_i}$ of the graph ${\mathscr G_i}$ and then combined in one correspondence set $\mathscr M_{\mathcal C_i}$ for the collection $\mathcal C_i$, as in Eqn. \ref{eq_cc}.
\begin{eqnarray}
\mathscr M_{\mathcal C_i} = \bigcup_{\forall_{e \in \mathscr E_i}} \mathscr M_{j,k}(\mathcal N_j, \mathcal N_k), \; \mbox{where} \; \{\mathcal N_j, \mathcal N_k \} = e. \label{eq_cc}
\end{eqnarray} 
The pairs in the set $\mathscr M_{\mathcal C_i}$, when each is considered as a graph edge, form a new graph $\mathscr G_{\mathcal C_i}$ with multiple connected sub-graphs $\mathscr G_{\mathcal C_i, s}$, where $s=1,\dots,S_i$. 
\begin{eqnarray}
\mathscr G_{\mathcal C_i} &=& (\mathscr V_{\mathcal C_i}, \mathscr E_{\mathcal C_i}). \\
\mathscr G_{\mathcal C_i, s} &=& (\mathscr V_{\mathcal C_i, s}, \mathscr E_{\mathcal C_i,s}), \; \mbox{} \;
{\tiny \mathscr V_{\mathcal C_i, s} \subset \mathscr V_{\mathcal C_i} \wedge  \mathscr E_{\mathcal C_i,s} \subset \mathscr E_{\mathcal C_i}}.  \label{eq_subg}
\end{eqnarray} 
The number of the connected sub-graphs $S_i$ ideally should equal to the number of the descriptors per ensemble and the number of vertices of each subgraph $|\mathscr V_{\mathcal C_i, s}|$ should equal to the number of the ensembles, $|\mathcal C_i|$, in the collection. However, this does not necessarily hold in practice. In fact, the number of the vertices can be less than $|\mathcal C_i|$ for some sub-graphs, due to possible failure to detect some key-points, or it can be particularly much higher than $|\mathcal C_i|$, since more than one sub-graph can be joined together due to mis-correspondences. When the number of the vertices of any sub-graph is significantly low, it can be considered as an indication that the underling key-point is not repeatable and the sub-graph is discarded. 
%The connected sub-graphs with comparable number of vertices to $|\mathcal C_i|$ are retained as is. 

The larger connected sub-graphs are iteratively segmented (partitioned) based on the well-known spectral graph partitioning (clustering) algorithm \cite{fiedler1973algebraic,donath1973lower}. The edges of the sub-graph to be partitioned are assigned the weight values shown in Eqn. \ref{eq_edgew}, where $w_{e_l}$ is the weight of the edge $e_l$ after normalization (by division by the average non-normalized weight $\bar w$). The distance $\bf d$ between the adjacent descriptors to the edge is the same defined in Eqn. \ref{eq_l1}.  
\begin{eqnarray}
w'_{e_l} = \frac{1}{{\bf d}(D_j, D_k)}, \; &&\mbox{where} \; \{D_j, D_k\} = e_l, \nonumber   \\ &&\mbox{for} \; l=1,\dots,|\mathscr E_{\mathcal C_i},s|. \\
w_{e_l} = \frac{w'_{e_l}}{\bar w}, \;\;\;\;\;\;\;\;\;\;\;\; &&\mbox{for} \; l=1,\dots,|\mathscr E_{\mathcal C_i},s|. \label{eq_edgew}
\end{eqnarray}
The Laplacian matrix ${\bf L}_{{\mathcal C}_i, s}$ of the weighted and connected sub-graph is then computed.
\begin{eqnarray}
{\bf L}_{\mathcal C_i, s} = {\bf D}_{\mathcal C_i, s} - {\bf A}_{\mathcal C_i, s}, \label{eq_lap}
\end{eqnarray}
where the diagonal matrix ${\bf D}_{\mathcal C_i, s}$ is the degree matrix of the weighted graph, each diagonal element is the sum of the normalized weights of all the incident edges to the corresponding vertex of the graph. ${\bf A}_{\mathcal C_i, s}$ is the adjacency matrix of the graph, each $a_{i,j}$ element is the normalized weight of the edge connecting the $i$-th vertex to the $j$-th vertex.
The second smallest eigenvalue $\lambda_f$ of ${\bf L}_{{\mathcal C}_i, s}$ is an indicator of how the graph is well-connected.
\begin{eqnarray}
{\bf L}_{\mathcal C_i, s} {\bf x}_{f} = \lambda_f {\bf x}_{f}. \label{eq_fv}
\end{eqnarray}
The corresponding eigenvector to $\lambda_f$ is known as the Fiedler vector, ${\bf x}_{f}$. The Fielder vector elements has comparable values for strongly connected vertices. In the next step, the K-means clustering algorithm (with only two clusters) is applied to the elements of ${\bf x}_f$. Based on the resulting two clusters of vertices, the graph ${\mathscr G}_{\mathcal C_i, s}$ partitioned. The described graph partitioning is iteratively applied until, the resulting graphs are strongly connected and roughly of the expected number of vertices, $|\mathcal C_i|$.    

Finally, the vertices (the descriptors) of the resulting connected graphs are considered as the equivalence sets $\mathcal Q_{\mathcal C_i,s}$, where $s=1,\dots, S_i$, of the collection $\mathcal C_{i}$. Each graph ${\mathscr G}_{\mathcal C_i, s}$ is then simplified to the star graph ${\mathscr T}_{\mathcal C_i, s}$ (the spanning graph).  Every vertex descriptor in the equivalence set is connected to one neutral descriptor which is chosen as the nearest to the means of $\mathcal Q_{\mathcal C_i,s}$ in case there are multiple neutral descriptors, called the bridging vertex (or descriptor), denoted by $B_{i,s}$. Similarly, the equivalence sets and the equivalence star graphs are extracted for every collection. The equivalence graphs are then allowed to join corresponding ones in all other collections, from the bridging to the bridging vertices. These joining edges represent the IRs. It is possible not to define any particular bridging points and allow for IRs connections (edges) from all the equivalence vertices of one collection to all the vertices of the corresponding sets in all other collections. 
%, according to the discussions in Section \ref{sec_ca}. 
In this case, the spanning graph of the equivalence sets will be a line connecting them, rather than a star. However, the definition of the bridging points significantly simplifies the graph search problem (the matching), as will be discussed in the next Section \ref{sec_gs}.            

\subsection{The Heuristic Graph Search}
\label{sec_gs}

The graph search takes place during the matching of a probe ensemble $\mathcal N_p$ to a gallery ensemble $\mathcal N_g$. The two ensembles are first corresponded as described in Section \ref{sec_cor}. Theoretically, an invariant dissimilarity measure can be computed between $\mathcal N_p$ and $\mathcal N_g$ for an optimal graph path $\mathcal P^* \in \mathscr P$ connecting them. %, as shown in Eqn. \ref{eq_ops}. %$\math N_p$ and $\mathscr N_g$.
In the general case, possible candidate paths start with $\mathcal N_p$ then zero or more collections are visited and finally terminate at $\mathcal N_g$. The paths should be simple, each has no repeated edges or collections.  At each visited collection two ensembles are visited (an entrance and an exit ones). At the descriptor level, there are multiple paths (a path bundle) connecting the corresponded descriptors in parallel for each higher level path. The paths are eventually realized as sequences of descriptors. The entrance and exit ensembles may be fixed for the descriptor level path bundle. However, relaxing this requirement and letting the entrance and the exit ensembles to vary for the different descriptor level paths is beneficial when the variety of the training expressions in the collections are limited. Below are the definitions of the paths at the different levels.     
\begin{align}
{\mathscr P'} &= \{\{\mathcal N_p, \mathcal C_{I(1)}, \dots, \mathcal C_{I(|\mathcal P|-2)} ,  \mathcal N_g\} \; | \; {\mathcal C_{I(.)} \in \mathscr C}\}.  \label{eq_pa} \\
&\equiv    \{\{\mathcal N_p, \mathcal N_{I(1),E(1)}, \mathcal N_{I(1),X(2)}, \dots, \nonumber %\\ &
\mathcal N_{I(|\mathcal P|-2), E(|\mathcal P|-2)},\mathcal N_{I(|\mathcal P|-2),X(|\mathcal P|-2)}, \mathcal N_g\} \; | \; \nonumber %\\&
{\mathcal N_{I(.), E(.)}} \in {\mathcal N} \wedge {\mathcal N_{I(.), X(.)}} \in {\mathscr N}\}. \\ 
&\equiv    \{\{D_{p,i}, D_{I(1),E(1),Q(2)}, D_{I(1),X(2),Q(2)}, \dots, \nonumber %\\&
D_{I(|\mathcal P|-2), E(|\mathcal P|-2), Q(|\mathcal P|-2)}, D_{I(|\mathcal P|-2),X(|\mathcal P|-2), Q(|\mathcal P|-2)}, \nonumber %\\&
D_{g,j}\} \\& \;\; | \; %\nonumber \\&
{D_{I(.), E(.), Q(.)}} \in {\mathscr D} \wedge {D_{I(.), X(.) Q(.)}} \in {\mathscr D} %\nonumber \\&
\wedge \{D_{p,i}, D_{g,j}\} \in \mathscr M_{p,g} \} = \mathscr P.
\end{align}
The multiple subscript indices of the path vertices uniquely point to the specific graph vertices and also indicate their identity $I(.)$, expression whether it is the entrance expression $E(.)$ or the exit expression $X(.)$, and the equivalence set $Q(.)$. 
Eqn. \ref{eq_ops} shows how the invariant measure can be computed based on the graph paths.
\begin{align}
s' =& \min_{\mathcal P \in \mathscr P} \sum_{\{{\bf x}, {\bf y}\} \in \mathscr M_{p,g}} \| \mathcal E_{X(|\mathcal P|-2)}^{g} \circ \mathcal I_{I(|\mathcal P|-2)}^{g} \dots \mathcal E_{E(1)}^{X(1)} \circ \nonumber \\ &
\mathcal I_{p}^{I(1)}({\bf x}) - {\bf y} - \Delta \mathcal E_p^g \| \\
  =& \min_{\mathcal P \in \mathscr P} \sum_{\{{\bf x}, {\bf y}\} \in \mathscr M_{p,g}} \| \mathcal I_{I(|\mathcal P|-2)}^{g} \circ \dots \circ \mathcal I_{p}^{I(1)}(\bf x) - {\bf y}\|. \label{eq_ops}
\end{align}
Existing optimal path searching algorithms such the Dijkstra and Bellman-Ford are not suitable for the solution of Eqn. \ref{eq_ops}. The Dijkstra algorithm deals with positive scalar edge weights while the the Bellman-Ford algorithm can also handle negative weights for more time complexity, $O(|\mathscr V|.|\mathscr E|)$, in comparison to $O(|\mathscr V|^2)$. In contrast, the edge weights in the tackled problem here are multidimensional vectors and the optimized quantity is the norm of their summation. It would be possible to devise a new variant of the Bellman-Ford algorithm that can handle this problem for possibly even more time complexity. However, it is impractical for the large and dense (with many edges) graph considered in the proposed approach.             

By considering only the bridging points, which were described earlier in Section \ref{sec_cg}, as the entrance and the exit vertices between collections, the graph density reduces and the maximum number of collections per path also reduces to two. This means that each considered path has at maximum three identity changes and four expression changes. Apart from the constraint on the maximum path length, $|\mathcal P|$, Eqn. \ref{eq_ops} holds for this lighter version of the graph search problem.

The proposed heuristic graph search proceeds by assigning one collection to the probe ensemble and one collection to the gallery ensemble (possibly another one). These assignments are initially performed based on the vicinity (in the descriptor space) of the descriptors of the probe and the gallery ensembles to those in the assigned collections. A KD-tree of all the training descriptors was built during an off-line stage to enable an efficient search for the nearest neighbors. A table of the descriptors information containing their ensemble, collection and equivalence sets is associated with the KD-tree. The $k$ nearest neighbors of each descriptor in the probe or the gallery ensemble vote for the different collections based on their associated information (labels). The collection which receives the highest number of votes is assigned to the ensemble. Next, the descriptors of $\mathcal N_p$ and $\mathcal N_g$ are assigned entrance and exit descriptors within the assigned collections. For this task, a separate KD-tree per collection was built (during an off-line stage), since smaller KD-tree are more efficient to search. Then, for each corresponded descriptor pair, $\{{\bf x}, {\bf y}\} \in \mathscr M_{p,g}$  where ${\bf x} \in \mathcal N_p$ and ${\bf y} \in \mathcal N_g$, the nearest three descriptors to $\bf x$ are considered as potential entrance descriptors to the collection assigned to $\mathcal N_p$. Similarly, the three nearest neighbors to $\bf y$ are found and considered as potential exit descriptors from the collection assigned to $\mathcal N_g$. Among the few combinations of potential entrance and exit descriptors, the one which yields the lowest value of the scalar function $m'({\bf x}, {\bf y})$, defined in  Eqn. \ref{eq_mval}, is assigned to $\bf x$ and $\bf y$ as respectively their entrance and exit descriptors.        
\begin{align}
m'({\bf x}, {\bf y}) = \| \mathcal I_{I(2)}^{g} \circ \mathcal I_{I(1)}^{I(2)} \circ \mathcal I_{p}^{I(1)}(\bf x) - {\bf y}\|. \label{eq_mval}
\end{align}
At this point of time, only a good initial guess of the solution is found and the search for the optimal path (or measure) is not performed yet. Nonetheless, the most similar people (collections) and expressions are likely to be assigned to the probe and the gallery.

The optimization is also carried out implicitly as nearest neighbor search. The descriptors of $\mathcal N_p$ are first mapped (or displaced) as in Eqn. \ref{eq_dmap} which produces a new image of the gallery descriptors, ${\bf x}'_i$ for $i=1,\dots,|\mathcal N_p|$. The new descriptors are then used to re-assign the gallery both a new training collection and new entrance and exit descriptors as described earlier. It is then followed by a similar mapping and re-assignment of the probe descriptors based on the image of the gallery descriptors, ${\bf y}'_i$ as in Eqn. \ref{eq_dmap2}. Each of the two steps implies the minimization of Eqn. \ref{eq_mval}.      
\begin{align}
{\bf x}'_i =  \mathcal I_{I(2)}^{g} \circ \mathcal I_{I(1)}^{I(2)} \circ \mathcal I_{p}^{I(1)}({\bf x}_i). \label{eq_dmap}\\
{\bf y}'_i =  \mathcal I_{I(1)}^{p} \circ \mathcal I_{I(2)}^{I(1)} \circ \mathcal I_{g}^{I(2)}({\bf y}_i). \label{eq_dmap2}
\end{align}
This process is then iterated a few times. In the early iterations, the re-assignments include re-assignment to new collections. Once the assigned collections converge, the later iterations involve only re-assignments to the entrance and the exit descriptors within the converged collections. The re-assignments to the entrance and the exit descriptors are only committed when they result in further minimization of Eqn. 
\ref{eq_mval}.          

The described graph search accounts for paths with two and one collection (as both the probe and the gallery can be assigned to the same collection). The direct path between the probe and the gallery should also be considered which is accounted for by the simple minimization in Eqn. \ref{eq_mval2}. 
\begin{align}
m({\bf x}, {\bf y}) = \min \{m'({\bf x}, {\bf y}), \|{\bf x} - {\bf y}\| \}. \label{eq_mval2}
\end{align}

\subsection{The Dissimilarity Measure between Ensembles}

An overall dissimilarity measure, $s$, between any two ensembles can be computed from the dissimilarity measures between the corresponded descriptor pairs, i.e., the $m$ values shown in Eqn. \ref{eq_mval2}. The $N$ descriptor measures with the least values are simply summed to produce the overall ensemble measure $s$, as in Eqn. \ref{eq_dissim_m}, where $N$ is much less than the typical number of the corresponded descriptor pairs. This would avoid the measures with high values that were not sufficiently reduced by the proposed matching approach, i.e., those without a sufficient level of expression removal. When computing the dissimilarity matrix, its entries are further normalized to range from zero to one for each probe. 
\begin{align}
s({\mathcal N_p}, {\mathcal N_g}) = \sum_i^N m_i \label{eq_dissim_m}
\end{align}

\section{Experiments}
\label{sec_exp}

A number of face recognition experiments were conducted on the FRGC v2.0, the 3D TEC and the Bosphorus datasets. These datasets differ from each other in a number of aspects. First, the FRGC dataset has the largest number of individuals (466 people in the testing partition) among the three datasets. It has diverse facial expressions but about half of its facial scans are under neutral or near neutral expressions. On the other hand, the 3D TEC and the Bosphorus datasets can even pose a more significant challenge to the recognition under facial expression variations. In the case of the 3D TEC, the challenge mainly arises because the individuals are identical twins (107 twins/ 214 individuals) and for the third and the fourth 3D TEC experiments the twins probes and the galleries are under different facial expressions. In contrast, the probe and the gallery scans specified in the dataset for the first and the second 3D TEC experiments involves no expression variations. In the case of the Bosphorus dataset, there are many scans for only 105 different people. However, the facial expression challenge arises because the facial expressions are generally of a larger extent in comparison to the other two datasets. 

As the proposed expression invariant approach requires a set of training data with many individuals and under different facial expressions including the neutral expression, the FRGC dataset is an appropriate choice for training the system. The FRGC dataset has a training partition. However, it has a limited number individuals and the individual sets of the training and the testing partitions are not mutually exclusive. For this reason, a significant part of the testing partition of the FRGC dataset was used for training the proposed system, all the facial scans of the first 300 individuals. The remaining scans of the testing partition was used for testing the proposed approach. A gallery of 166 neutral facial scans (one scan per subject) was formed. The remaining scans were split into a neutral and a non-neutral probe subsets. The trained system was then used to perform the tests on the 3D TEC and the Bosphorus datasets. In all the experiments including both the expression-invariant approach and the plain RAIK approach which was used for result comparison, the RAIK features \footnote{The RAIK descriptor has two adjustable parameters $\alpha$ and $\beta$ which were respectively set to -0.15 and 0.0 for all the conducted experiments.} were compressed using the principal component analysis (PCA), each to a vector of twenty PCA coefficients.

\begin{center}
\begin{figure}
\begin{tabular}{c @{\hspace{1mm}}c}
\includegraphics[width=0.4\columnwidth, clip=true]{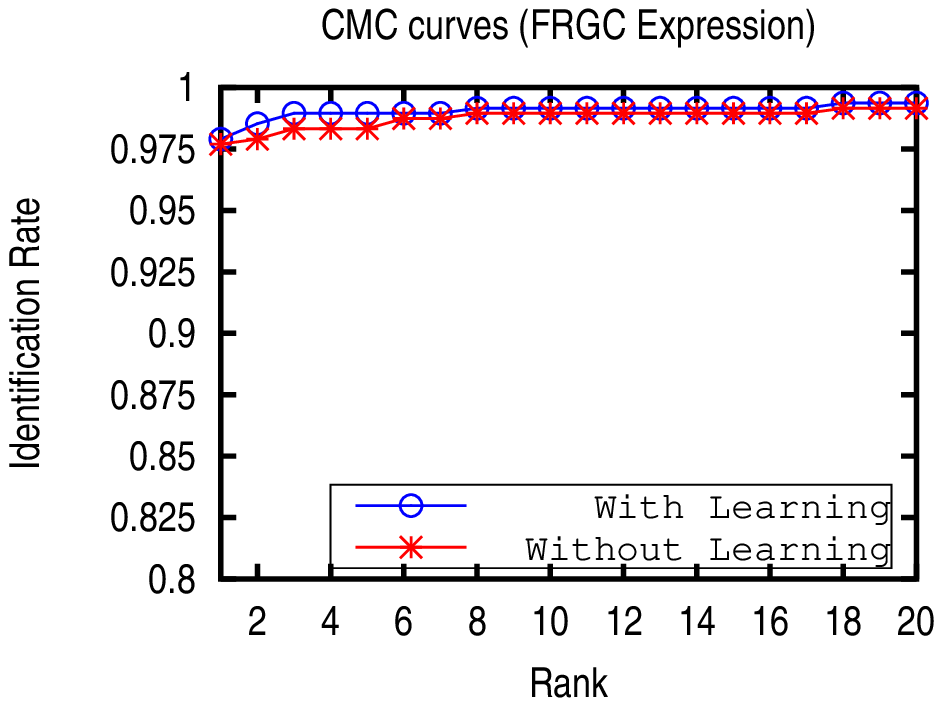} & \includegraphics[width=0.4\columnwidth, clip=true]{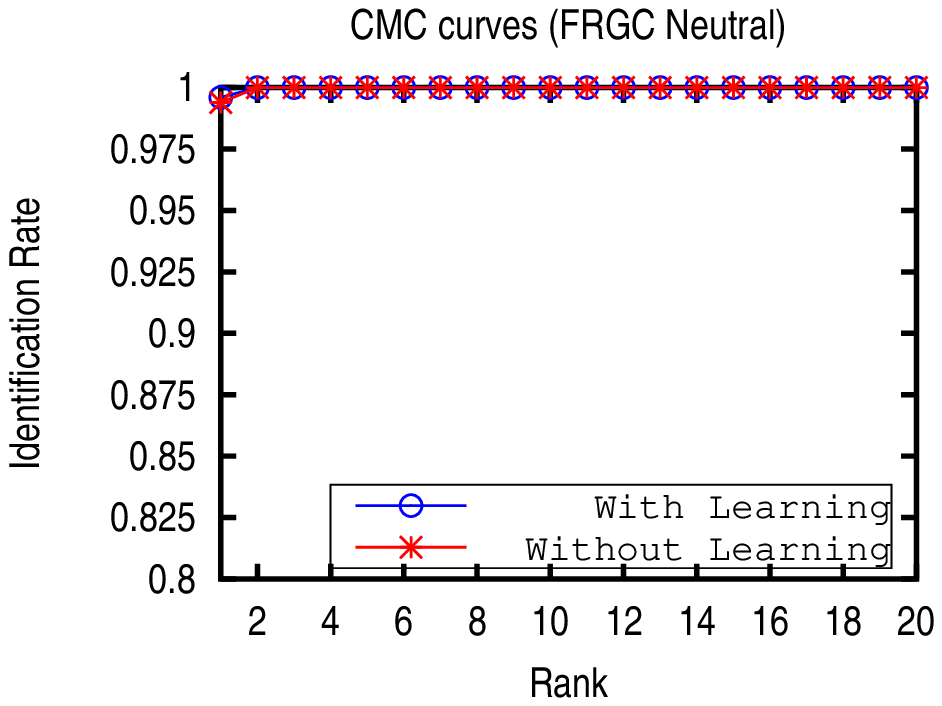}
%\textbf{text} & text
\end{tabular}
\vspace{0mm}
\caption{The CMC curves of the non-neutral (left) and the neutral (right) subsets of the FRGC dataset which were spared for evaluation. The curves compare the performance when using the proposed learning approach based on the RAIK features to that when the RAIK features are used without learning. Both methods achieve a very high performance. However, a noticeable improvement is observed for the non-neutral scans in particular, especially for the first few ranks.}
\label{cmc_frgc}
\end{figure}
\end{center}

\begin{center}
\begin{figure}
\begin{tabular}{c @{\hspace{1mm}}c}
\includegraphics[width=0.4\columnwidth, clip=true]{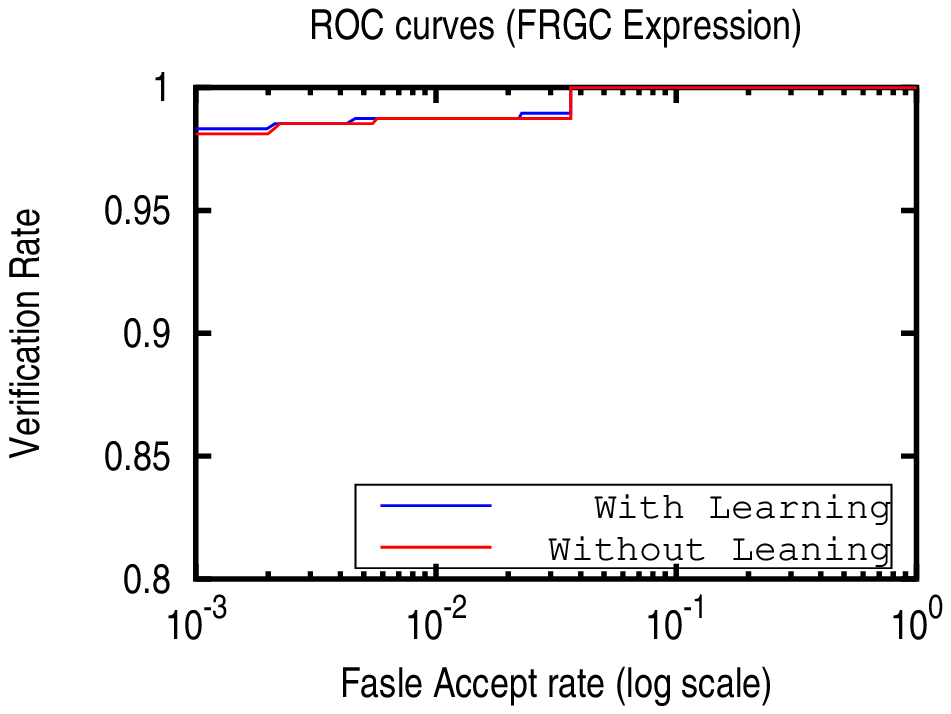} & \includegraphics[width=0.4\columnwidth, clip=true]{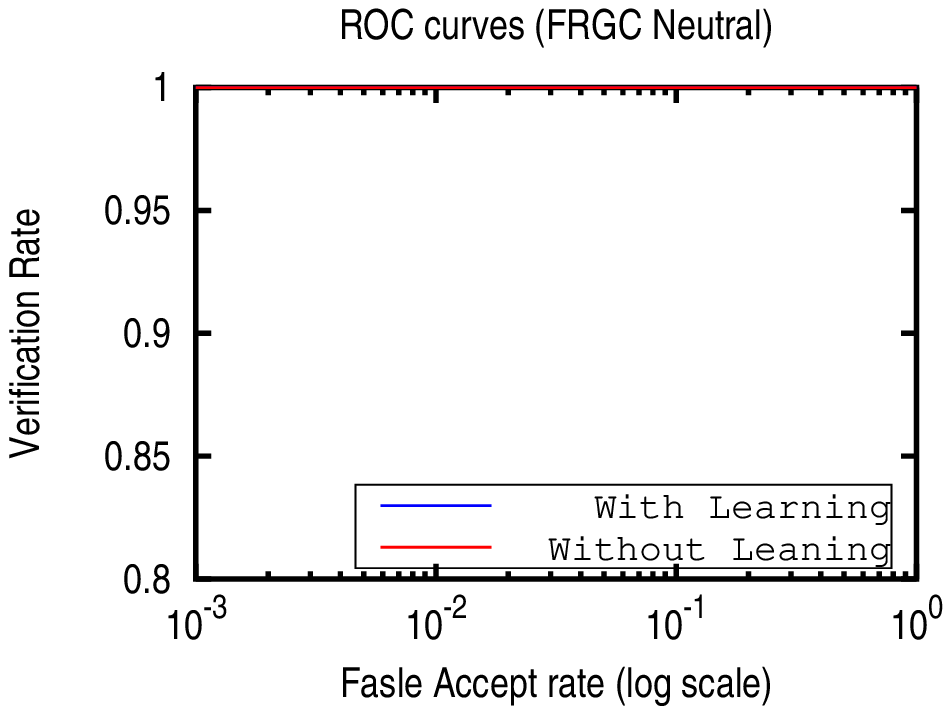}
%\textbf{text} & text
\end{tabular}
\vspace{0mm}
\caption{The ROC curves for the non-neutral and neutral subsets of the FRGC dataset which were spared for evaluation.}
\label{roc_frgc}
\end{figure}
\end{center}

The recognition performance results of the experiments conducted on the FRGC dataset, Fig. \ref{cmc_frgc} and \ref{roc_frgc}, indicate that the proposed expression-invariant approach noticeably enhances the identification rates of the non-neutral probes at the first few ranks where the expression variations have more impact on the identification performance. While the first rank identification rate has increased from $97.69\%$ (for the plain RAIK approach) to $97.90\%$ (for the proposed apporach based on the RAIK descriptors), the margin between the two rates has further increased at the second rank and peaked at the third rank where the identification rate has increased from $98.32\%$ to $98.95\%$. It should be noted that the plain RAIK approach already achieves a very high identification performance because it limits the matching to the semi-rigid regions (the forehead and the nose) of the face. As more regions are considered, the identification rate margin between the proposed approach and the plain RAIK approach increases in favor of the proposed approach. This is because the identification rates of the plain RAIK approach declines more rapidly with the inclusion of the non-rigid regions of the face while proposed approach still declines but a slower pace. However, the performance of both approaches is optimal when only the semi-rigid regions of the face are considered. It could be concluded that the proposed approach contributes the reliability of the recognition in addition to the observed recognition performance enhancement. For the neutral experiment, the impact of the proposed approach is limited, which is expected as there are no expression variations and the identification rates for the neutral scans are already above $99.5\%$. Some verification rate improvement was observed for the non-neutral experiment but it was not significant. It has increased from $98.11\%$ to $98.31\%$ at 0.001 FAR.  

The identification and verification rates of the first and the second experiments of the 3D TEC dataset were not significantly impacted by the proposed approach. The interpretation of this observation is that for these two experiments the probe and the gallery scans of the twins are under the same expressions. In contrast, the impact of the proposed approach on the third and the fourth experiments was more significant. The proposed approach has increased the first rank identification rate of the third experiment from $85.51\%$ to $89.25\%$ and from $86.45\%$ to $89.25\%$ for the fourth experiment. It appears from the 3D TEC and the FRGC results that the recognition enhancement of the proposed approach becomes more significant when the expression variations are more challenging to the plain RAIK approach. For these two experiments, the verification rates were respectively $92.52\%$ and $91.12\%$ at 0.001 FAR for the proposed approach, in comparison to $88.79\%$ and $88.32\%$ at 0.001 FAR for the plain RAIK approach. The results of the Bosphorus dataset indicate that the proposed approach enhances the recognition performance for the probes under non-neutral facial expressions. The identification rate had increased from $91.94\%$ to $93.55\%$ and the verification rate had increased from $92.60\%$ to $94.21\%$ at 0.001 FAR. There was a negligible degradation in the verification performance of the neutral expression scans. Nonetheless, both the proposed system and the plain RAIK system had achieved above $99.5\%$ verification at 0.001 FAR for the neutral scans.   

\begin{center} 
\begin{figure}
\begin{tabular}{c @{\hspace{1mm}}c}
\includegraphics[width=0.4\columnwidth, clip=true]{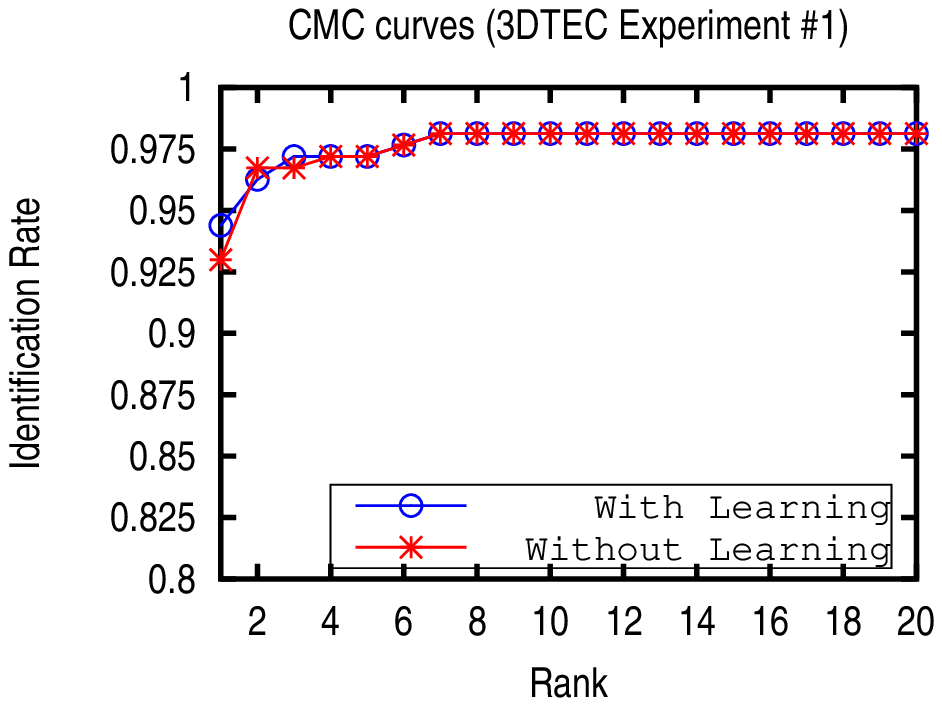} & \includegraphics[width=0.4\columnwidth, clip=true]{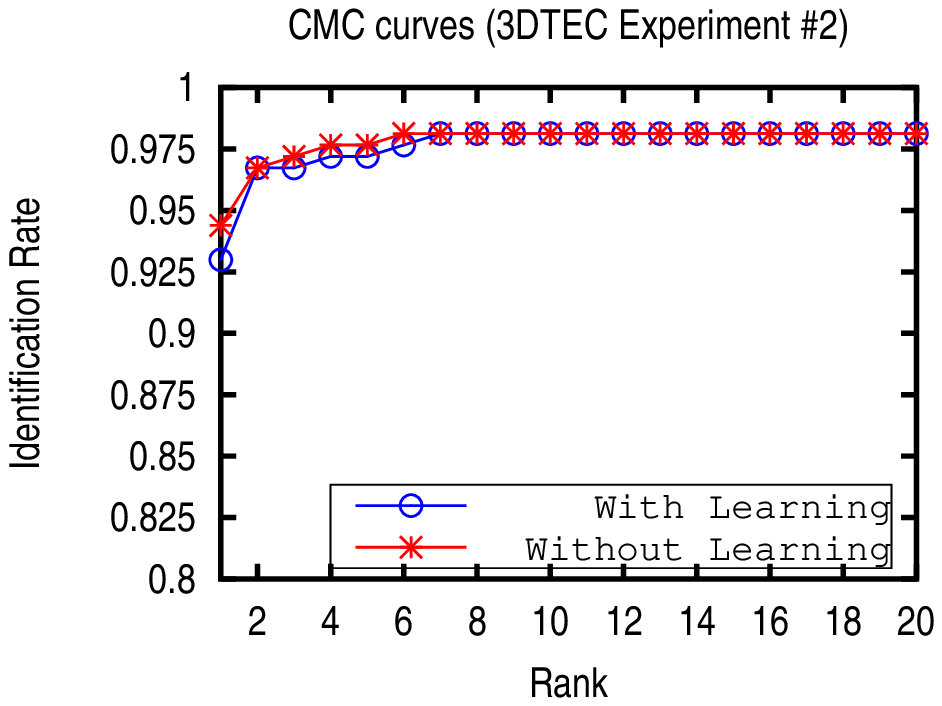} \\
\includegraphics[width=0.4\columnwidth, clip=true]{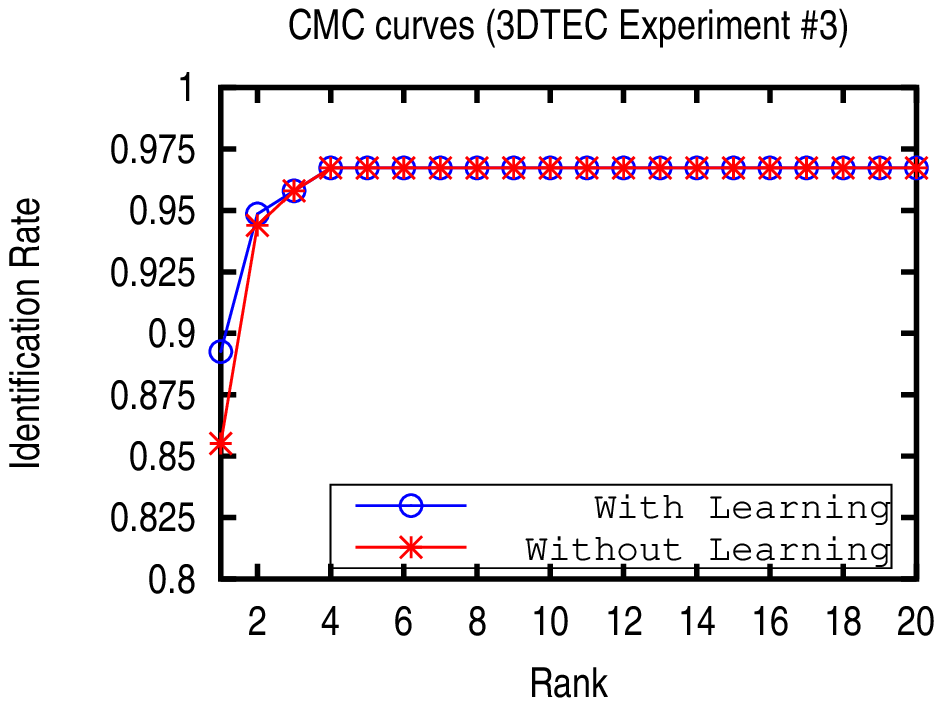} & \includegraphics[width=0.4\columnwidth, clip=true]{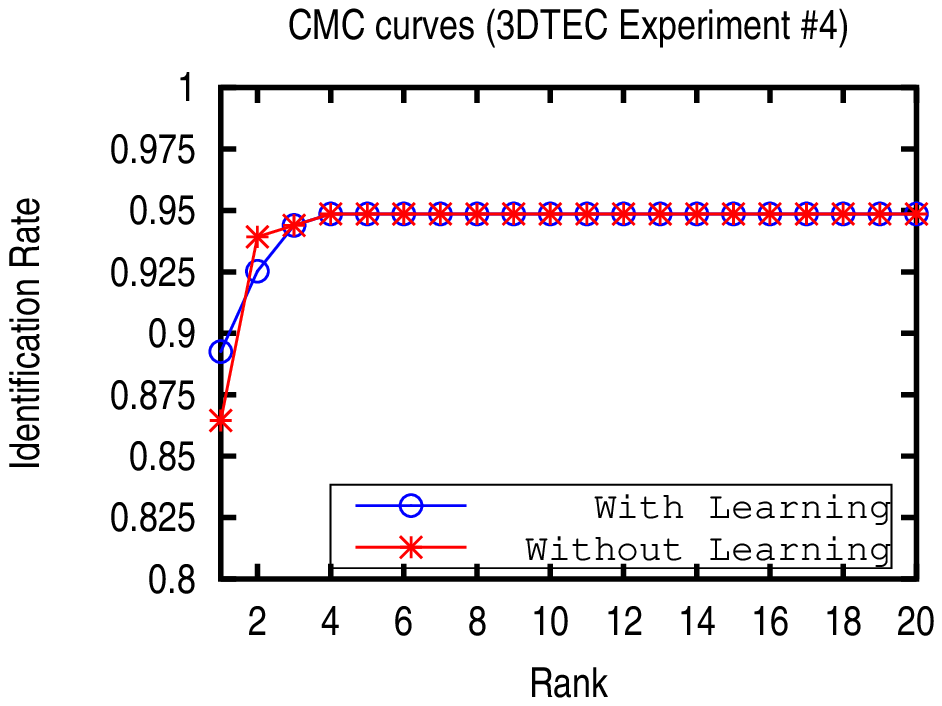}
%\textbf{text} & text
\end{tabular}
\vspace{-3mm}
\caption{The CMC curves of the four experiments of the 3D TEC dataset. The first two experiments involve no expression variations. For these two experiments there is a limited effect of the learning approach on the identification performance. In contrast, the third and fourth experiments have expression variations among the twins facial scans, for which the proposed approach has shown to considerably improve the recognition performance.}
\label{cmc_3dtec}
\end{figure}
\end{center}
\begin{center}
\begin{figure}
\begin{tabular}{c @{\hspace{1mm}}c}
\includegraphics[width=0.4\columnwidth, clip=true]{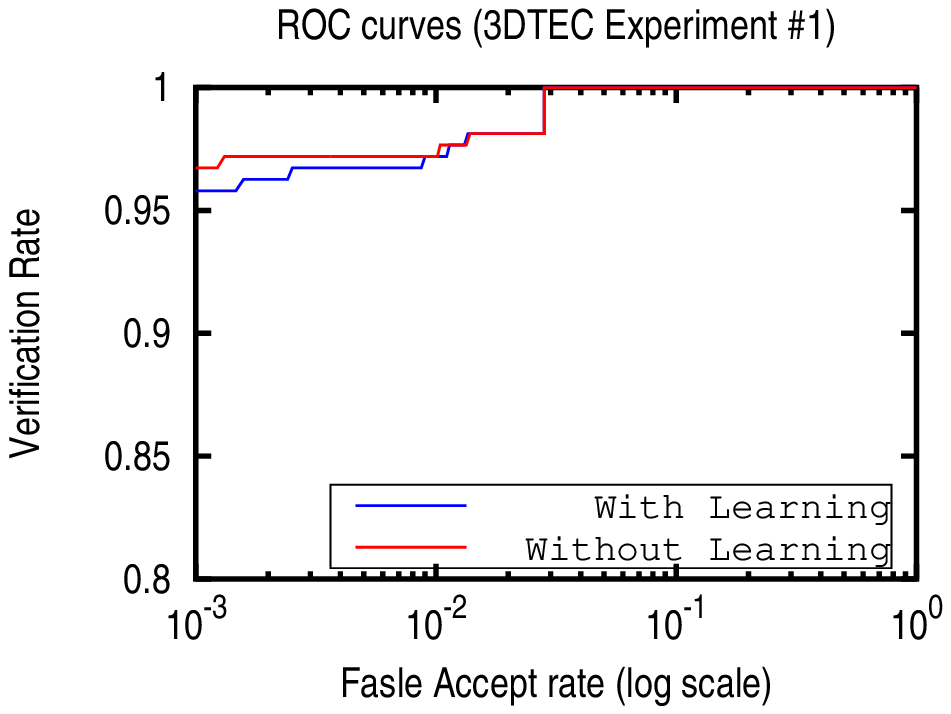} & \includegraphics[width=0.4\columnwidth, clip=true]{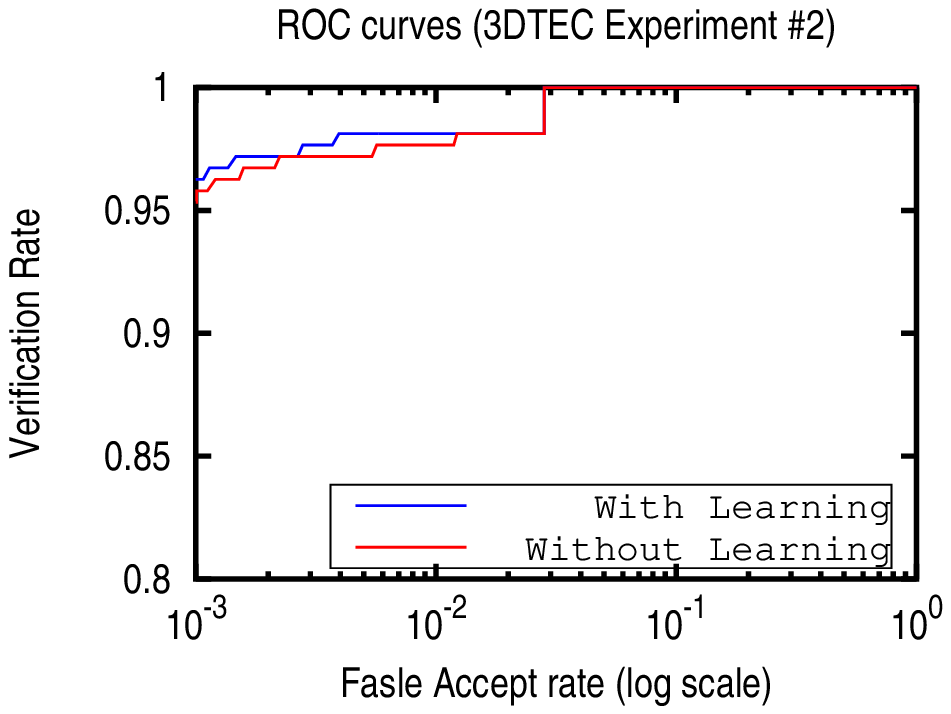} \\
\includegraphics[width=0.4\columnwidth, clip=true]{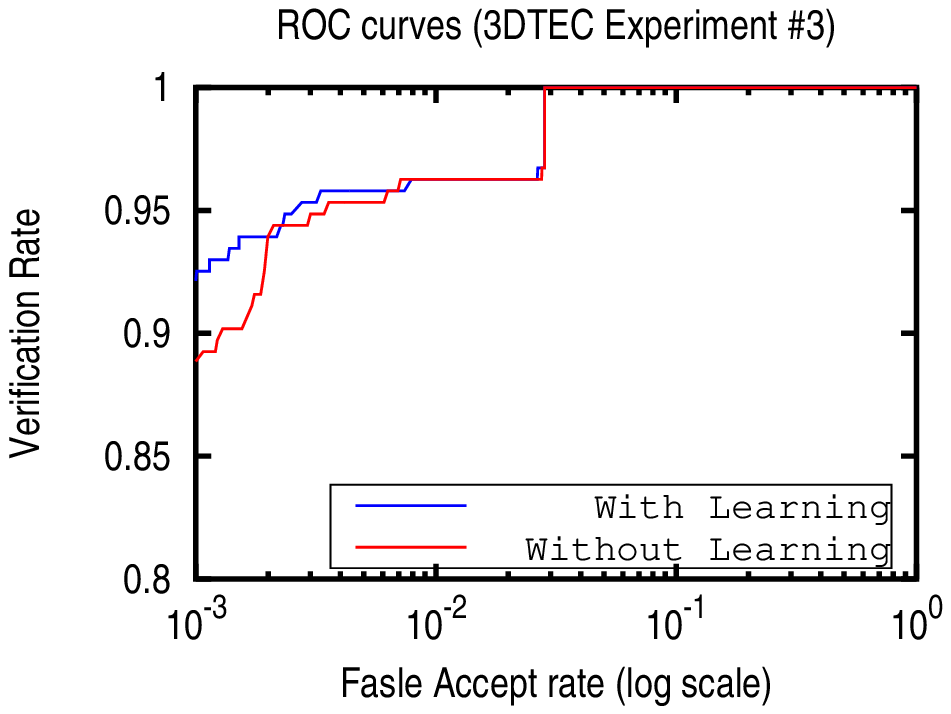} & \includegraphics[width=0.4\columnwidth, clip=true]{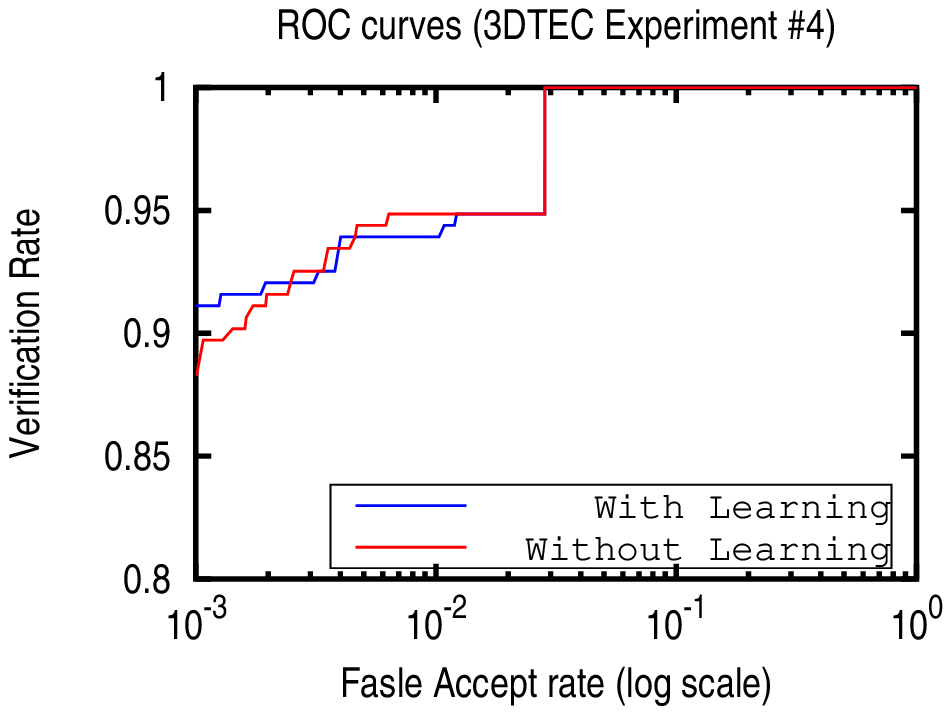}
%\textbf{text} & text
\end{tabular}
\vspace{-3mm}
\caption{The ROC curves of the four 3D TEC experiments.}
\label{roc_3dtec}
\end{figure}
\end{center}
\begin{center}
\begin{figure}
\begin{tabular}{c @{\hspace{1mm}}c}
\includegraphics[width=0.4\columnwidth, clip=true]{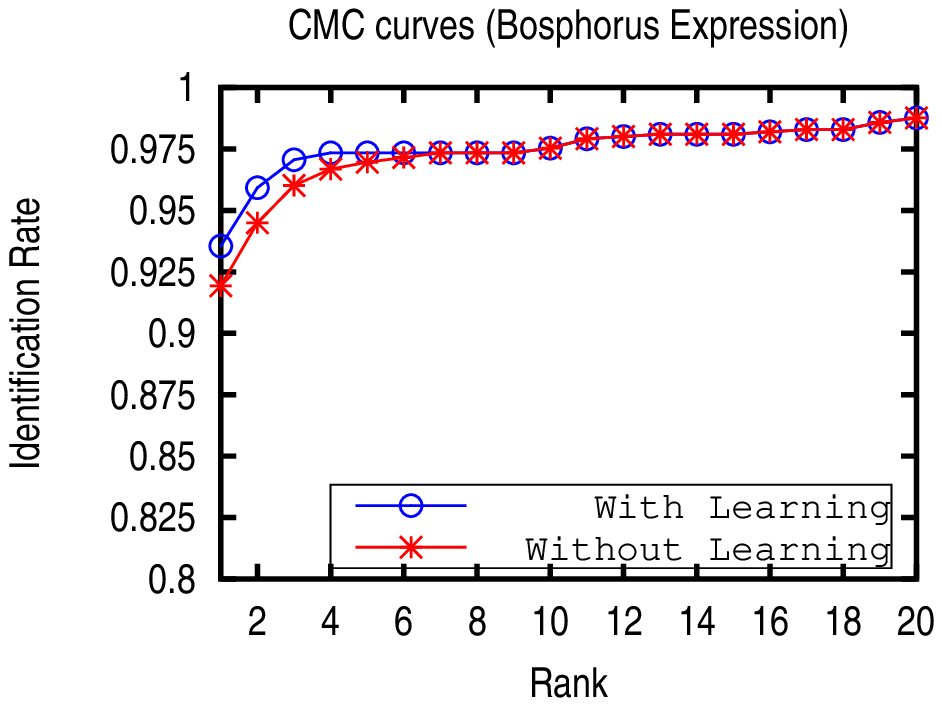} & \includegraphics[width=0.4\columnwidth, clip=true]{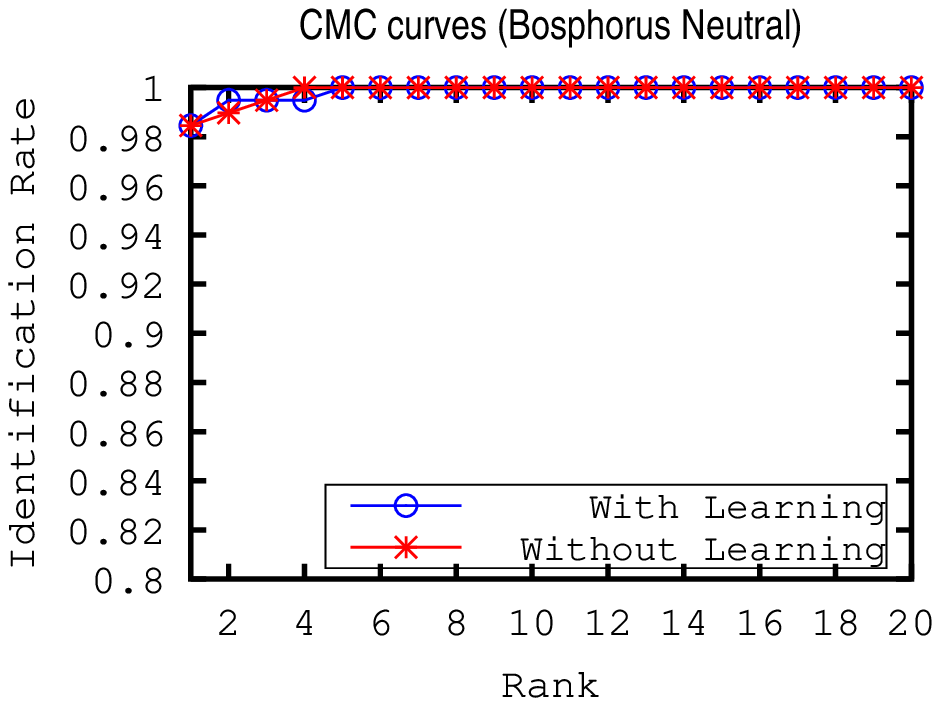}
%\textbf{text} & text
\end{tabular}
\vspace{0mm}
\caption{The CMC curves of the Bosphorus dataset.}
\label{cmc_bosph}
\end{figure}
\end{center}
\begin{center}
\begin{figure}
\begin{tabular}{c @{\hspace{1mm}}c}
\includegraphics[width=0.4\columnwidth, clip=true]{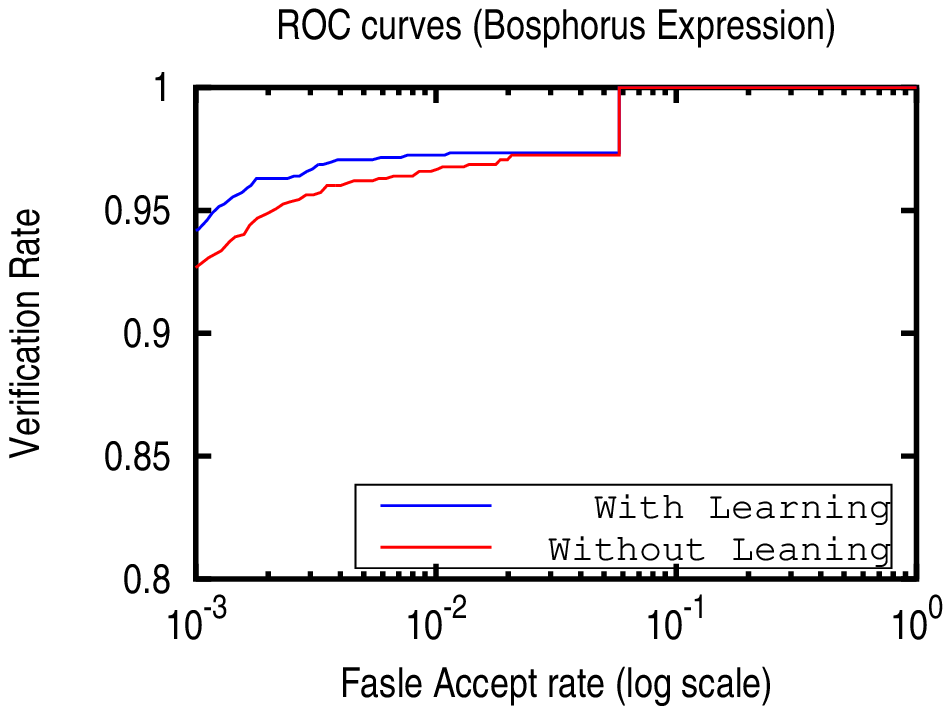} & \includegraphics[width=0.4\columnwidth, clip=true]{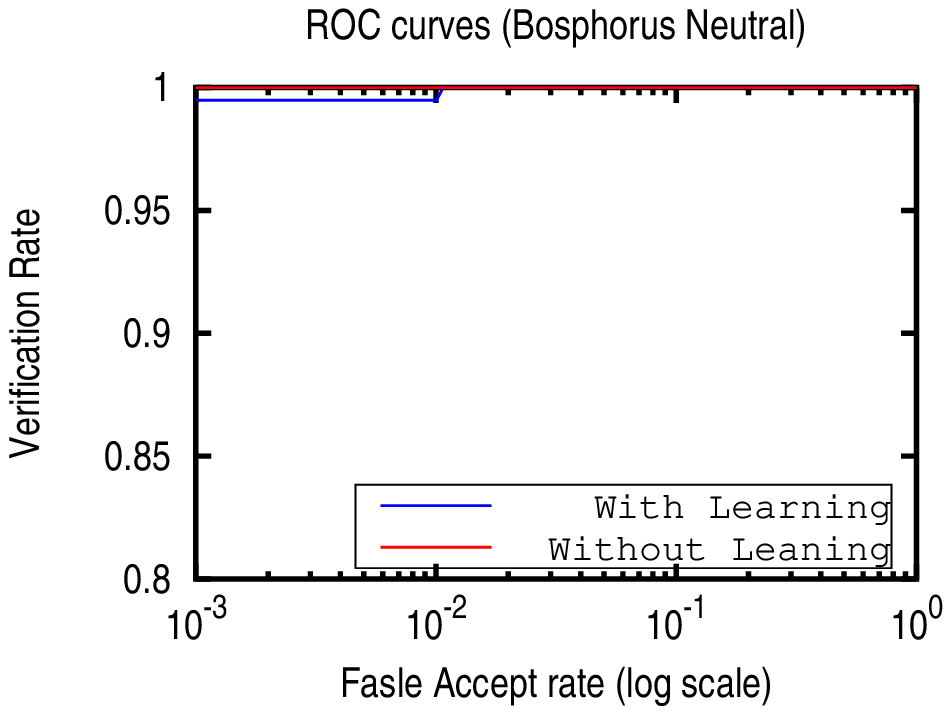}
%\textbf{text} & text
\end{tabular}
\vspace{0mm}
\caption{The ROC curves of the Bosphorus dataset.}
\label{roc_bosph}
\end{figure}
\end{center}

\section{Conclusion}
\label{sec_conc}

The proposed research described a new manifold-based approach for learning facial expression invariance of the key-point descriptors. The descriptor variations induced by the facial expressions were handled through the use of the equivalence relations within the descriptor manifold. Then, invariant dissimilarity measures (distances) were computed based on the equivalence relations. This approach has shown to improve the recognition performance when the facial scans being matched involve expression variations.   



%% References


\bibliographystyle{elsarticle-num}
\bibliography{myrefs}






% \end{thebibliography}

\end{document}